\documentclass{article}
\usepackage{spconf,amsmath,epsfig}
\usepackage[ruled,vlined]{algorithm2e}
\usepackage{algorithmic}
\usepackage{multirow}
\RequirePackage{booktabs} 
\usepackage{color,soul}   

\usepackage{tabularx}
\usepackage{graphicx}
\usepackage{adjustbox}

\title{Sample selection for efficient image annotation}
\twoauthors
{Bishwo Adhikari, Esa Rahtu}{Tampere University, Finland}
{Heikki Huttunen}{Visy Oy, Finland}

\begin{document}

\maketitle
\begin{abstract}
Supervised object detection has been proven to be successful in many benchmark datasets achieving human-level performances. However, acquiring a large amount of labeled image samples for supervised detection training is tedious, time-consuming, and costly. In this paper, we propose an efficient image selection approach that samples the most informative images from the unlabeled dataset and utilizes human-machine collaboration in an iterative train-annotate loop. Image features are extracted by the CNN network followed by the similarity score calculation, Euclidean distance. Unlabeled images are then sampled into different approaches based on the similarity score. The proposed approach is straightforward, simple and sampling takes place prior to the network training. Experiments on datasets show that our method can reduce up to 80\% of manual annotation workload, compared to full manual labeling setting, and performs better than random sampling. 

\end{abstract}
%
\section{Introduction}
\label{sec:intro}

Object detection is one of the fundamental and widely studied problems in computer vision. The success of convolutional neural networks (CNN) and deep learning has transformed the domain of object detection. These approaches outperform earlier techniques by a large margin, but still behind the human-level understanding. Recently, object detection has been mature enough to be used in applications in various domains such as agriculture~\cite{object_detection_in_agriculture}, medical imaging~\cite{Lutnick_2019}, robotics~\cite{takuya_robotics}, and remote sensing~\cite{remote_sensing_2020}.

Supervised object detection is the most widely used approach but requires a large amount of labeled examples for the training, with the labels usually assigned by human annotators. In object detection, the datasets consist of images, and each image can have multiple annotations; the acquisition of a large number of high-quality datasets is tedious, time-consuming, costly, and often requires expertise (e.g. medical images). Object detection breakthroughs in various fields have been facilitated by a large number of annotated benchmark datasets available from multiple domains~\cite{pascal_voc_2007, kitti_dataset, coco_dataset}. Object detection tasks can be very specific to a certain environment and condition, thus relying only on public datasets often does not generalize well. 

Several existing works have been reported on methods for data-efficient object detection training including transfer learning~\cite{transfer_learning}, semi-supervised learning~\cite{Self_Semi_iccv_2019}, and weakly supervised learning~\cite{weakly_supervised_2018}. However, most of these methods still require a certain amount of domain-specific labeled training data.
Crowd-sourcing~\cite{crowdsourcing} have been increasingly popular due to the progress in third-party services such as Amazon Mechanical Turk (AMT) for data annotation. As there is no standard to measure the quality of the annotations, the annotated datasets via these platforms often have noisy labels \textit{i.e.}, missing labels, inaccurate labels, and only part of the object is being labeled. Moreover, not all projects can opt for these solutions due to the privacy concern and cost of high-quality annotation from expert annotators.
Active learning~\cite{settles_active_learning} is another popular technique that aims to reduce the training time by actively querying for the labels of unlabeled instances. Recently, active learning techniques have been commonly used for the image classification task but not commonly available for the object detection task. Hence, there is a need for faster and efficient methods for labeling image examples for object detection model training. 

In this work, we focus on reducing manual annotation workload using active image sampling. Our method selects and sorts informative images into mini-batches that will lead to less annotation work; utilizing the human-machine collaboration in the continuous train-annotate loop. We use a distance-based image sampling based on the entire image-to-image pairwise Euclidean distance. We propose two sampling approaches that can reduce the overall annotation workload. Secondly, we utilize a self-supervised network for the feature extraction.
Experiments on three datasets show that the proposed method significantly reduces the annotation workload compared to the full human-level manual annotation. The saving in annotation cost is up to 80\% in the best case and above 50\% on average.

The rest of the paper is structured as follows.
The literature review of related works is presented in Section~\ref{sec:experiment}. In section~\ref{sec:method}, we present a detailed description of our method and its components. In Section~\ref{sec:experiment}, details about the experimental setups, dataset, and results are presented and discussed. Finally, we conclude this paper with our findings in Section~\ref{sec:conclusion}. 

\vspace*{-\baselineskip}
\begin{figure*}[ht]
    \centerline{\includegraphics[scale=1.2]{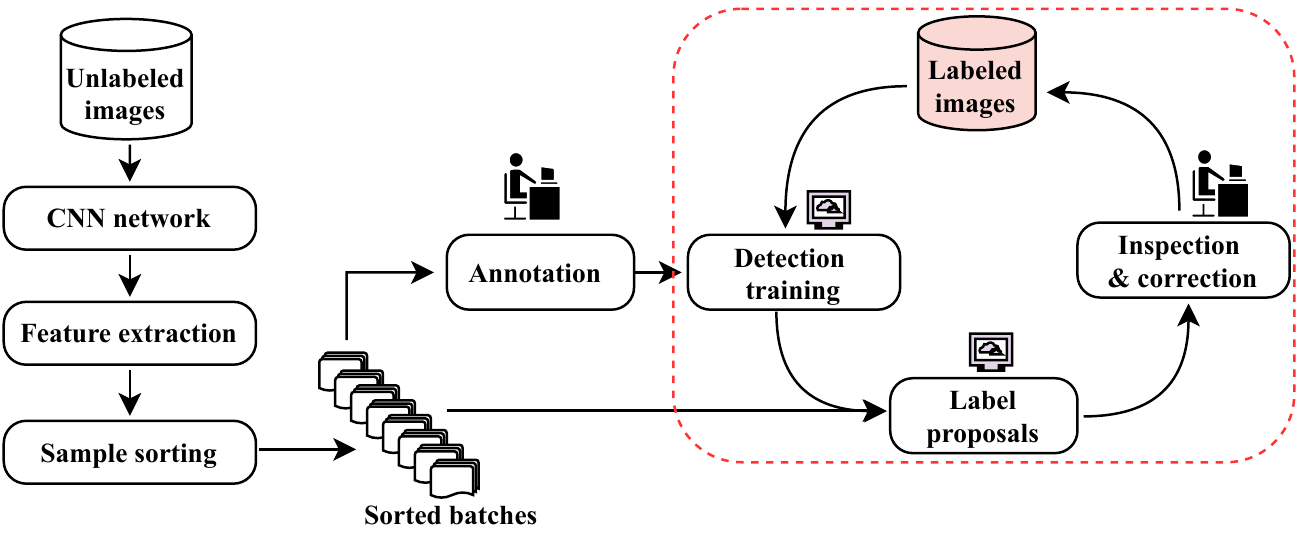}}
    \caption{Proposed efficient annotation scheme. The sample selection is done before the object detection model training. The train-annotation loop (inside red) continues till all unlabeled batches are taken and labeled.}
\label{fig:method}
\end{figure*}
\section{Related Work} 
\label{sec:related_work} 

\subsection{Active Learning}
\label{ssec:active}
Active learning~\cite{settles_active_learning} aims to reduce the labeling cost by selecting the most useful samples from the unlabeled dataset and request for the labels to maximize a model performance while trained with these labels. Active learning has been proven to be effective in reducing annotation cost on image classification task~\cite{Li_CVPR_2013}.

For object detection, uncertainty-based sampling is commonly used as a measure of informativeness~\cite{Desai_bmvc_2019,Yoo_2019_CVPR}. The key principle of this approach is to train a learner, calculate the degree of uncertainty, and then query an unlabeled instance with the least confidence.
The drawback of uncertainty-based approaches is that they rely on a labeled dataset to build an initial model to select the query instance, and the performance is often unstable when there are only a few labeled samples available.
Querying every unlabeled sample using a CNN network is computationally expensive. In~\cite{MedAL_2018}, a feature descriptor function was proposed to generate features for all images and the most distant sample is selected based on the euclidean distance measure from these features. This is feasible for a single-class dataset but requires heavy computation since every sampling instance requires a new round of training and evaluation with the CNN model.

Our method takes all unlabeled images for the sampling prior to network training, representing all samples from the unlabeled dataset. Furthermore, it does not rely on the prediction of classification scores, hence it is efficient and straightforward. The proposed sample selection method is close to~\cite{MedAL_2018} but in this work we sampled all images based on the distance metric before the network training, resulting in a computationally efficient workflow.

\subsection{Self-training}
\label{ssec:Straining}
Many previous works have used a trained model to predict labels for a set of unlabeled images and involve humans in the process of correcting predicted labels~\cite{Lutnick_2019,learning_intelligent_dialogs,adhikari_2018,Assistive_2019,adhikari_2020}.
In~\cite{adhikari_2018}, a two-stage method proposed for speeding up bounding box annotation; model training on a small set of a labeled dataset, inference labels for the unlabeled set followed by manual correction on network proposals. In~\cite{adhikari_2020}, an iterative train-annotate loop is proposed for efficient image annotation in a small batch of images at a time. This method explicitly used the information from ground truth for the selection of images on three approaches.
In~\cite{Assistive_2019},  assistive learning feedback loop is proposed that utilizes the contextual sampling criteria; uniqueness, and average euclidean distance of the images. However, in reality, often such information is not available for the unlabeled dataset. In this work, we do not need such information from unlabeled samples.

The major difference from all these works is that (1) our method does not require explicit information from the unlabeled images; (2) all the sampling task is done before the training phase; (3) we use an entire image for the sample selection based on the extracted features; and (4) human annotator has a rather easy task of labels inspection and/or correction that does not need prior machine learning skills. Among these works, the iterative annotation approach~\cite{adhikari_2020} is the most similar to the proposed one.

\section{Method}
\label{sec:method}
The proposed method applies CNN based feature extractor to sample the images in mini-batches based on the distance metric \textit{i.e.,} most similar or most distant images together. Human annotator labels first batch ($B_1$), trains object detection model ($M_1$) on this batch $B_1$. The freshly trained model $M_1$ is used to predict bounding boxes and class labels on the unlabeled batch ($B_2$). The predicted bounding boxes and labels are then inspected and corrected by a human annotator resulting in a fully labeled batch $B_2$. The next iteration starts with training model $M_2$ on a combination of labeled $B_1$ and $B_2$ and predicts labels for the batch $B_3$.
The train-prediction-correction loop continues till the last batch $B_n$ of unlabeled images is labeled. A human annotator is required for the manual labeling of  $B_1$ and in the inspection and correction stage, inside the loop, as shown in Figure~\ref{fig:method}.

\subsection{Feature Extraction}
\label{ssec:feature}
The first step in the proposed method is to extract image features from all images in the unlabeled dataset. For the feature extraction, we experiment with two networks; ResNet50~\cite{Resnet_Kaiming_He} network trained on ImageNet~\cite{imagenet} (ImgNet) and similarity network (SimNet) trained on images from the unlabeled dataset. Moreover, any CNN network can be utilized as a feature extraction network.
SimNet uses ResNet50 as a backbone and structured lifted loss~\cite{struct_loss} as loss function. Images from the unlabeled dataset are horizontally split into pairs, and the SimNet is trained on these pairs for 25 epochs. Experiments showed that features extracted from this network perform better in sample selection \textit{i.e.,} mapping similar and dissimilar samples.

\subsection{Image Sampling}
\label{ssec:sampling}
The extracted image features are used to compute the pairwise Euclidean distances of each pair. The images are then sampled based on the distance measure and sorted into batches, later used in the iterative train-annotate loop. We have experimented with three strategies for batch selection; similar (images with the least distance are placed together), dissimilar (images with the maximum distance are placed together), and random sampling (images are randomly sampled into batches). Additionally, we experiment on temporal order if applicable to the dataset.

We start with randomly selecting few images as a query list. On a dissimilar approach, an image that has the largest distance compared to all images in the query list is appended to the query list one at a time, the process continues till all images are added to the list, resulting in distinct images together. While, on a similar approach, an image that has the least distance compared to all images in the query list is added to the list one at a time till all images are ordered in the list, resulting in similar images together.

\subsection{Training, Inference and Correction}
\label{ssec:training}
At first, the detection network is trained on the manually labeled first batch $B_1$. After the first round, it is trained on all labeled sets available at that time. The process continues till all unlabeled samples are being used in the proposal stage or the desired amount of data is labeled. The number of iterations and other hyperparameters can be decided depending on the available annotation budget.

The freshly trained model is used to infer bounding box and class labels on the batch of unlabeled images. After this, the task of inspection and correction on these proposals is assigned to a human annotator. This task includes adding missing bounding boxes and labels, correcting incorrectly drawn boxes and labels, and removing extra boxes and labels. As the model gets better, the more tedious task is done by the machine, thus simplifying the task of a human annotator.

\subsection{Workload Calculation}
\label{ssec:calculation}
We follow the previous works~\cite{adhikari_2018, adhikari_2020, Desai_bmvc_2019} in formulating the annotation workload calculation. However, unlike~\cite{adhikari_2020}, the annotation costs include the workload required to label images from the first batch $B1$, and costs are assumed to be different for different actions during the correction stage.

The annotation workload in terms of bounding boxes is given as
\[\text{Workload ($W_B$)} = \text{\# corrections} + \text{\# of object in $B_1$}\]
where \# corrections represents the sum of additions ($A_n$) and removals ($R_n$) from the manual correction stage. Simply counting the number of added and removed bounding boxes.

On average, these addition and removal actions, required at the correction stage, have different levels of complexity. Hence, the number of objects (bounding boxes) does not provide a realistic estimate of the entire dataset labeling cost. Therefore, we formulate the labeling cost in terms of time (sec.) as
\[\text{Workload ($W_T$) } = \text{ $B_{1n}\times T$} + \text{$A_n\times T$} + \text{$R_n\times T/2$}\]
where $B_{1n}$, $A_{n}$, and $R_n$ are the number of bounding boxes in $B_1$ (labeled by the annotator at the beginning), manual additions and removals respectively. While $T$ is the time to annotate one object from scratch.

In~\cite{adhikari_2018}, the annotation workload calculation uses only the manual correction stage. However, in practice, taking into account all manual work required from all stages would give a better estimation of the annotation workload. The time estimation to draw bounding boxes is conservative and there have been various estimations presented in~\cite{learning_intelligent_dialogs, adhikari_2018, Assistive_2019}. In our work, we calculate the annotation cost reduction (\%) compared to what it would have taken in a full manual annotation.

\section{Experiments}
\label{sec:experiment}

\subsection{Datasets}
\label{ssec:dataset}
\textbf{\em PASCAL VOC --- }%
PASCAL VOC 2007~\cite{pascal_voc_2007} is extensively used in vision research. The object detection version is fully labeled in 20 object classes including animals, vehicles, and common household objects. In our experiments, we use all 9963 images with 30638 object instances from both trainval and test sets. In this dataset, we performed two sets of experiments: first with all class categories and second with individual class categories.

\textbf{\em KITTI --- }%
KITTI~\cite{kitti_dataset} object detection and object orientation estimation benchmark consist of 7481 training images and 7518 test images, comprising a total of 80256 labeled objects. The dataset is collected in five scenarios: city, residential, road, campus, and person. We use the train split from object detection category consisting of 7481 images with 40570 object instances of 8 class categories including vehicles (car, van, tram, truck), and person involved in different actions (pedestrian, sitting, cyclist).

\textbf{\em Indoor --- }%
Indoor dataset~\cite{adhikari_2018} is a moderate size dataset collected from indoor premises. This dataset has 2213 images and about 4595 object instances of 7 indoor scene classes. The dataset consists of safety signs (exit, fire extinguisher), furniture(trash bin, chair), and equipment (clock, printer, screen). Images were extracted from a series of videos preserving temporal order.

\subsection{Implementation Details}
\label{ssec:implement}
For object detection, we experiment on two commonly used detection networks: single-stage SSD~\cite{liu2016ssd} with MobileNetV2~\cite{sandler2018mobilenetv2} backbone and two-stage Faster RCNN~\cite{ren2015faster} with ResNet50 backbone. Both networks are fine-tuned with pre-trained weight from COCO dataset~\cite{coco_dataset}. 
We select these networks following their popularity and speed vs accuracy trade-off reported in the literature. The SSD network is lighter and faster but less accurate than the alternatives. However, it is a common choice for resource-limited scenarios. On the other hand, Faster RCNN is more complex, computationally heavy but optimal for the high-quality annotation proposals. We resized images to  $300\times300$ pixels for the SSD network and $600\times1024$ pixels for the Faster RCNN network, as reported on their original papers.

\begin{figure}[ht]
    \centerline{\includegraphics[clip, width=\linewidth]{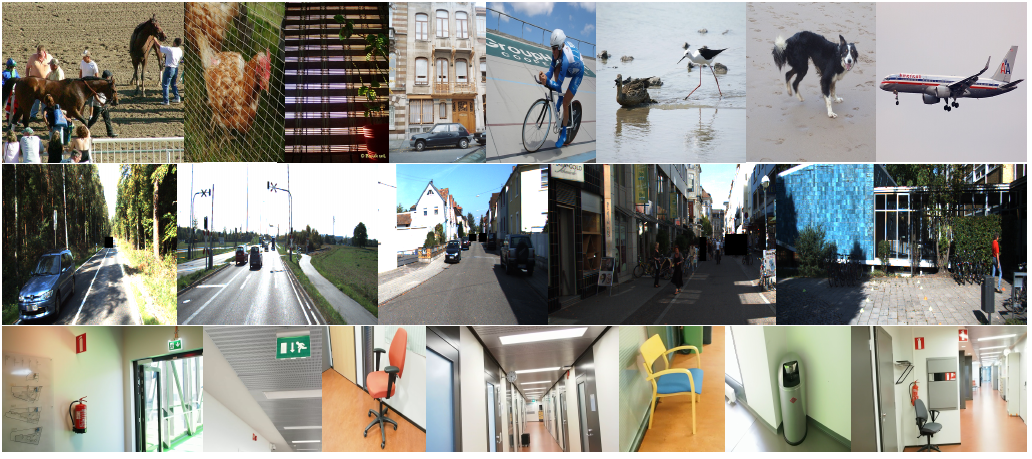}}
    \caption{Example samples selected using the dissimilar approach. Images from PASCAL VOC, KITTI, and Indoor datasets from top to bottom rows, respectively.}
\label{fig:example}
\end{figure}

We mainly experiment on three different sampling strategies: similar, dissimilar, and random shuffle, mentioned in~\ref{ssec:sampling}. An additional experiment is done with temporal sampling on the Indoor dataset. In Figure~\ref{fig:example}, we show images sampled by our dissimilar approach in three datasets.
\vspace*{-\baselineskip}
\begin{table}[ht]
    \caption{Reduction of manual annotation time $W_T$ (\%), higher the better, with different approaches. The result is an average of five independent runs. The best for each dataset per column is in bold font.}
    \label{table:result_time}
    \begin{tabular}{c|c|c|c||c}
    \hline
    \multirow{2}{*}{Dataset} & \multirow{2}{*}{Approach} & \multicolumn{2}{c||}{RCNN} & SSD \\
    \cline{3-5}
    \multirow{2}{*}{} & \multirow{2}{*}{} & 0.5 IoU & 0.7 IoU & 0.5 IoU \\
    \hline \hline
    & Shuffle & 56.00 & 55.37 & 31.61 \\ \cline{2-5}
    & Sim (ImgNet) & 56.05 & 54.71 & \textbf{32.97} \\
    Pascal & Dis (ImgNet) & 56.48  & 55.63 & 32.08 \\
    VOC & Sim (SimNet) & 55.27 & 53.65 & 30.53 \\
    & Dis (SimNet) & \textbf{56.82} & \textbf{56.02} & 32.08 \\
    \hline\hline
    & Shuffle & 50.82 & 49.53  & 32.21 \\ \cline{2-5}
    & Sim (ImgNet) & 37.31 & 34.68 & 19.92 \\
    KITTI & Dis (ImgNet) & \textbf{51.49}  & \textbf{50.59} & \textbf{33.61} \\
     & Sim (SimNet) & 46.67 & 43.34 & 28.64 \\
    & Dis (SimNet) & 49.81 & 47.16 & 28.89 \\
    \hline \hline
    & Shuffle & 81.20 & 80.37 & 65.98 \\ \cline{2-5}
    & Temporal & 37.47  & 35.99 & 20.24 \\
    & Sim (ImgNet) & 59.12 & 59.47 & 43.35 \\
    Indoor & Dis (ImgNet) & \textbf{81.83} & 78.81 & 64.09 \\
     & Sim (SimNet) & 67.06 & 67.38 & 48.32 \\
    & Dis (SimNet) & 79.08 & \textbf{81.64} & \textbf{67.76} \\
    \hline
\end{tabular}
\vspace{-4mm}
\end{table}

\vspace*{-\baselineskip}
\begin{table}[ht!]
    \caption{Reduction of manual workload (\%) in terms of bounding boxes $W_B$ with different approaches. The result is an average of five independent runs. The best for each dataset per column is in bold font.}
    \label{table:result_bbox}
    \begin{tabular}{c|c|c|c||c}
    \hline
    \multirow{2}{*}{Dataset} & \multirow{2}{*}{Approach} & \multicolumn{2}{c||}{RCNN} & SSD \\
    \cline{3-5}
    \multirow{2}{*}{} & \multirow{2}{*}{} & 0.5 IoU & 0.7 IoU & 0.5 IoU \\
    \hline \hline
    & Shuffle & \textbf{43.69} & 47.11 & 26.17 \\ \cline{2-5}
    & Sim (ImgNet) & 43.64 & 47.18 & \textbf{27.50} \\
    Pascal & Dis (ImgNet) & 43.12  & 47.39 & 26.46 \\
    VOC & Sim (SimNet) & 43.40 & 46.22 & 26.15 \\
    & Dis (SimNet) & 43.66 & \textbf{48.22} & 2.15 \\
    \hline\hline
    & Shuffle & 45.93 & 45.59  & 27.90 \\ \cline{2-5}
    & Sim (ImgNet) & 28.08 & 28.12 & 14.49 \\
    KITTI & Dis (ImgNet) & \textbf{44.39}  & \textbf{46.33} & \textbf{29.37} \\
     & Sim (SimNet) & 39.24 & 38.49 & 23.89 \\
    & Dis (SimNet) & 42.58 & 42.57 & 25.03 \\
    \hline \hline
    & Shuffle & 77.11 & 78.19 & 60.83 \\ \cline{2-5}
    & Temporal & 34.77  & 34.29 & 17.60 \\
    & Sim (ImgNet) & 56.10 & 57.25 & 39.43 \\
    Indoor & Dis (ImgNet) & \textbf{78.10} & 76.51 & 58.57 \\
     & Sim (SimNet) & 63.17 & 65.07 & 43.25 \\
    & Dis (SimNet) & 74.08 & \textbf{78.43} & \textbf{63.70} \\
    \hline
\end{tabular}
\end{table}

\vspace*{-\baselineskip}
\begin{table*}[ht]
\caption{Annotation workload reduction(\%) in time ($W_T$) and bounding boxes ($W_B$) on individual class categories of the PASCAL VOC 2007 dataset. The best result per category is in bold font.}
\label{table:results_voc}
\begin{adjustbox}{width=\textwidth}
\begin{tabular}{l|c|c|c|c|c|c|c|c|c|c|c|c|c|c|c|c|c|c|c|c||c}
\toprule
  Approach &\textbf{Plane} &\textbf{Bicycle} &\textbf{Bird} &\textbf{Boat} &\textbf{Bottle} &\textbf{Bus} 
  &\textbf{Car} &\textbf{Cat} &\textbf{Chair} &\textbf{Cow} &\textbf{Table} &\textbf{Dog}	
  &\textbf{Horse}	&\textbf{Bike} &\textbf{Person} &\textbf{Plant}	&\textbf{Sheep}	&\textbf{Sofa} 
  &\textbf{Train}	&\textbf{TV}  &\textbf{Average}\\
\midrule
Dis SimNet ($W_T$)   
& 41.12  & \textbf{57.00} & 44.12 & \textbf{24.39} & 40.97 & \textbf{41.30} & 67.30 & 58.82 & \textbf{48.25} 
& \textbf{41.38} & 37.35 & 62.68 & \textbf{58.80} & \textbf{58.98} & 74.59 & \textbf{38.78} 
& 28.39 & 45.13  & \textbf{52.69}  & 44.71 & 48.34 \\  
Dis SimNet ($W_B$)  
& 30.21 & \textbf{47.71} & 34.97 & \textbf{18.71} & 36.94 & \textbf{29.18}  & 59.15  & 46.11 & \textbf{37.91} 
& \textbf{36.05} & 11.49 & 49.71 & \textbf{51.06} & \textbf{49.01} & 66.54 & \textbf{29.91} 
& 25.30 & 17.54  & \textbf{41.58} & 36.53 & 37.78\\
\midrule
Shuffle ($W_T$)    
& \textbf{46.80} & 53.22 & \textbf{46.93} & 23.51 & \textbf{45.66} & 40.58 & \textbf{68.74} & \textbf{60.80} & 47.71 
& 40.95 & \textbf{41.70} & \textbf{64.27} & 54.74 & 51.91 & \textbf{75.54} & 37.83 
& \textbf{29.21} & \textbf{45.49}  & 49.44 & \textbf{47.94} & \textbf{48.65} \\ 
Shuffle  ($W_B$)    
& \textbf{39.40} & 45.10 & \textbf{39.57} & 16.68 & \textbf{40.58} & 27.85 & \textbf{59.65} & \textbf{47.43} & 37.24 
& 37.51 & \textbf{25.78} & \textbf{50.37} & 45.07 & 44.79 & \textbf{67.53} & 29.98 
& \textbf{26.65} & \textbf{22.17}  & 36.98   & \textbf{38.59} & \textbf{38.95} \\ 
\bottomrule
\end{tabular}
\end{adjustbox}
\vspace*{-\baselineskip}
\end{table*}

\subsection{Results}
\label{ssec:results}
We compare the performance of the proposed sampling approaches for labeling the full dataset in a simulated environment. The amount of workload reductions for different datasets with different sampling approaches and two detection networks is shown in Table~\ref{table:result_time}. Additionally, we experimented on two IoU thresholds; $0.5$ and $0.7$ which are commonly used as performance measurement in object detection. In all experiments, our sampling approaches perform better in saving manual annotation time. The Faster RCNN model with dissimilar sampling performs better in most cases. The difference between the two IoU thresholds is surprisingly small. We believe this is due to our calculation strategy. In an ideal case, with 0.5 IoU, the network proposes too many labels which the annotator needs to remove in the correction stage. While with 0.7 IoU, the network purposes fewer tight boxes where the annotator needs to add some labels manually, at the correction stage.

In~\cite{adhikari_2020}, the workload reduction is calculated in terms of the bounding boxes. We show the saving in terms of bounding boxes in Table~\ref{table:result_bbox}. However, we believe that saving in terms of time would be more practical. Next, we experiment on all categories of PASCAL VOC 2007 taking one class at a time, shown in Table~\ref{table:results_voc}. We trained a Faster RCNN network with two approaches: random shuffle and dissimilar sampling order based on our similarity network. 

\vspace*{-\baselineskip}
\begin{figure}[h!]
\centerline{\includegraphics[scale=0.71]{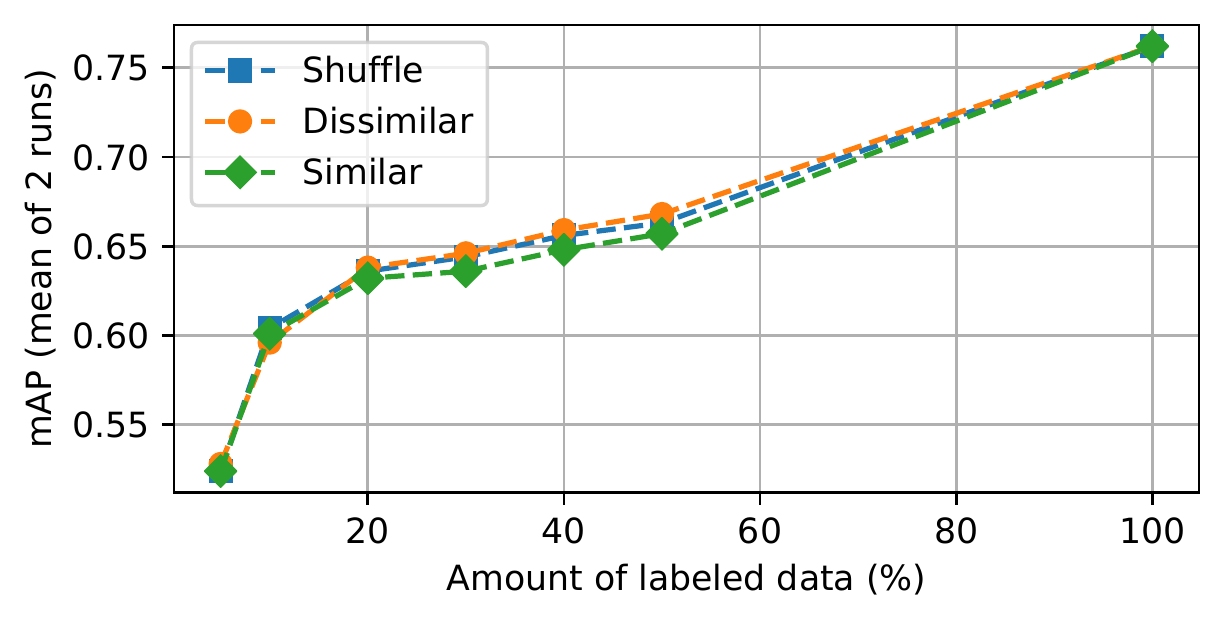}}
\vspace*{-\baselineskip}
\caption{Relationship between mean average precision (mAP) and amount of labeled dataset. Faster RCNN network is trained on PASCAL VOC 2007 \textit{all} with three sampling approaches and tested with PASCAL VOC 2012 \textit{trainval}.
}
\label{fig:annot_budget}
\end{figure}

Next, we experiment with the relationship between the amount of sampled data with the model performance on the PASCAL VOC 2012 \textit{trainval} set. The result in Figure~\ref{fig:annot_budget} indicates that the proposed methods improve the performance of the object detector when there are fewer labeled samples available for training. As the amount of labeled data increases the performance of the model given all three approaches converges.

\subsection{Discussion}
\label{ssec:discussion}
The reduction in workload varies among datasets. We believe this is due to the nature of datasets (e.g. type, size, and ground truth). For the Indoor dataset, which is extracted from video sequences and sparsely labeled, the saving is higher. For the KITTI dataset which is densely labeled, the saving is lower due to dense annotations. While in the PASCAL VOC dataset, which is collected from the web, manual workload reduction with our method is moderate. Since the dataset is already so diverse, there is less to achieve with sample selection.

Furthermore, the results obtained on a single class states that the annotation workload reduction with our method is higher for difficult classes. We get better results on classes such as potted plant, chair, and boat that are considered to be hard~\cite{img-difficulty-2016}. These difficult classes usually appear in complex contexts with many objects, occlusion, and varying illumination. In the case of a large-scale dataset even 1\% of workload reduction has a significant impact. For example, in the PASCAL VOC, 1\% of 30680 is 306 object instances. According to~\cite{crowdsourcing}, labeling this many objects takes about 3 hours.

The quality of labels obtained from the self-training method is equally important as the labeling cost. Since a human annotator is used to inspect and correct each object proposal in our method, the quality of the labeled dataset is good compared to other semi-supervised approaches. Experiment results with different IoU thresholds show that the proposed method is effective to get tighter bounding boxes, with 0.7 IoU threshold, the workload reduction in annotation time is higher than that of 0.5. This concludes that the proposed method is feasible for efficient sample selection for the cost-effective annotation campaign.

\section{Conclusion}
\label{sec:conclusion}
We proposed a similarity-based self-trained approach for efficient labeling of object detection datasets. The proposed approach can produce high-quality annotation with a reasonable annotation budget. Most of the tedious work is done by the machine while the human annotator mostly takes care of correction work which is often easier than labeling images from scratch. Extensive experiments on three datasets showed that a large amount of manual annotation work can be saved if some focus is paid on sample selection prior to the network training. In the future, we would like to apply this method to more challenging video datasets annotation and semantic annotation.

\end{document}